\documentclass[11pt]{article}

\usepackage[final]{acl}

\usepackage{times}
\usepackage{latexsym}
\usepackage[ragged]{footmisc}

\usepackage[T1]{fontenc}
\usepackage[utf8]{inputenc}

\usepackage{microtype}
\usepackage{amsmath}

\usepackage{inconsolata}
\usepackage{float}

\usepackage{graphicx}
\usepackage{tcolorbox}

\title{Entity tracking emerges in sub-billion parameter language models and exceeds human performance in naturalistic narratives}

\author{
  \textbf{Karolina Drożdż\textsuperscript{1}} \and
  \textbf{Micha Heilbron\textsuperscript{2,3}} \\
\\
  \textsuperscript{1}IDEAS Research Institute; Warsaw, Poland \\
    \textsuperscript{2}Max Planck Institute for Psycholinguistics; Nijmegen, Netherlands \\
  \textsuperscript{3}University of Amsterdam, Amsterdam Brain and Cognition; Amsterdam, Netherlands \\
  \texttt{karolina.drozdz@ideas.edu.pl}, \texttt{micha.heilbron@mpi.nl}
}

\usepackage{lipsum}
\usepackage{stfloats}
\begin{document}
\maketitle

\begin{figure*}[b]
  \centering
  \includegraphics[width=\textwidth]{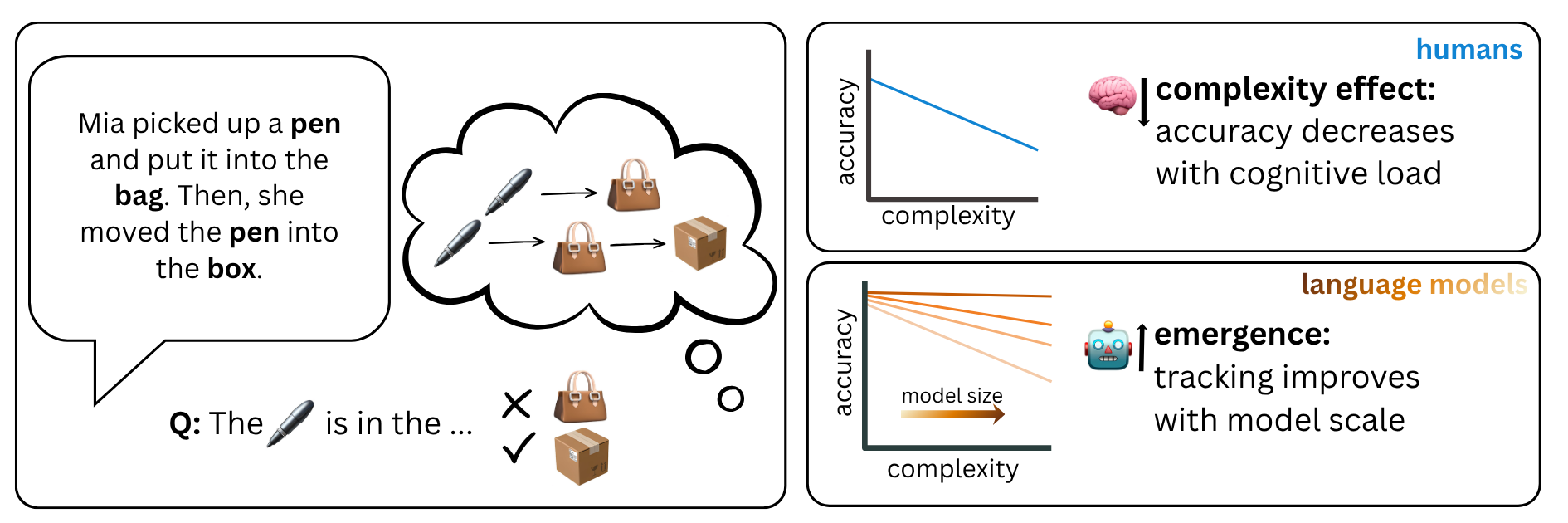}
  \caption{Conceptual overview. 
  Left: language understanding requires \emph{tracking entities} in an internal model of discourse context, as a narrative unfolds. 
  Right: in humans (top), entity tracking accuracy decreases with situational complexity; in language models (bottom), we find a similar pattern, and an increase in capability with scale.}
  \label{fig:conceptual_diagram}
\end{figure*}

\begin{abstract}
% \lipsum[1]
% TOPIC
Understanding language requires tracking entities across discourse -- i.e., knowing where things are and how they change, even when not explicitly stated.
% MOTIVATION
Whether language models perform such tracking in a human-like fashion remains unclear, in part because existing evaluations rely on artificial tasks, far removed from natural language comprehension, and lack comparisons to humans.
% CONTRIBUTION
Here, we evaluate entity tracking in both language models and humans ($N=48$) using naturalistic narratives at multiple levels of complexity. 
% We test two open model families across scales alongside two larger, contemporary models.
% Evidence/contribution 1
In humans, we find that entity tracking degrades specifically with narrative complexity, not narrative length. 
In language models, we find that human-level entity tracking is already present at 410 million parameters -- well below the multi-billion parameter, code-specialised models identified by prior work -- and improves with scale, with contemporary models far exceeding human performance. 
% EVIDENCE / contribution 2
% This capacity is present in base models, and is robust to pseudowords and semantically 
% improbable objects, indicating that models reliably track entities through discourse structure.
% IMPACT
Together, these results demonstrate that entity tracking, a core component of language understanding, emerges at model scales far smaller than previously thought.

\end{abstract}

\section{Introduction}

The fluency of large language models (LLMs) has reignited a long-standing question: does the capacity to produce coherent text imply genuine language understanding, or can it be mimicked through sophisticated pattern matching \citep{bender2020climbing, shanahan2024talking}? 
One way to operationalise this question is to ask whether language models (LMs) \emph{track entities} across discourse as human comprehenders do -- that is, maintaining and updating representations of the underlying states, locations, and relations of referents as a narrative unfolds \citep{GroenendijkStokhof1991,Heim2002,ZwaanRadvansky1998}. 
% Critically, this requires more than merely registering what is explicitly stated. 
Critically, many of these states are never explicitly mentioned but must be inferred from described events,  based on a coherent internal model of the text's meaning.
For instance, if a story describes a key being removed from a box containing only a key, comprehenders immediately infer that the box is now empty, even though this is never mentioned \citep{Lietal2021}. 
To what extent do language models that exhibit apparent understanding actually track entities, like a genuine comprehender?

Prior work has begun to address this question from several angles.
First, influential probing work showed that entity states could be linearly decoded from encoder-decoder model activations \citep{Lietal2021}, though subsequent work revealed that most of the reported accuracy derived from trivial cases rather than genuine entity tracking \citep{KimSchuster2023}.
Analysing model outputs, \citet{KimSchuster2023, Kimetal2024} found that robust entity tracking surfaced only in large models pretrained on massive, code-specialised datasets. % , rather than text alone. % , leading to the conclusion that code-inclusive pretraining is a critical factor. 
% -- though this comparison relied on closed models which have since been deprecated. 

%old
% - The linguistic fluency of large language models (LLMs) creates a compelling illusion of comprehension. However, generating coherent text does not guarantee that a model maintains an accurate internal representation of the world it describes. Assessing the \textit{formal} linguistic competence of LLMs (Mahowald et al., 2024) is now well established: controlled psycholinguistic paradigms show that models acquire generalizable syntactic knowledge without explicit supervision (Linzen, 2016; Hu et al., 2020; Linzen \& Baroni, 2021; Duan et al., 2024; Warstadt et al., 2020; Manning et al., 2020). In contrast, their \textit{functional} competence -- using language to represent and reason about the world -- remains comparatively underexplored, with few controlled paradigms analogous to those used for syntax (Mahowald et al., 2024). 

However, some limitations complicate the conclusions that can be drawn from this body of work. 
First, no study has compared model performance to human performance, despite evidence that humans routinely rely on shallow, ``good-enough'' strategies rather than building fully detailed situation models \citep{ferreira2002good}. 
The extent to which humans reliably track entities across narratives is thus an empirical question that matters for interpreting model performance. 
Second, the tasks used in prior evaluations are highly formulaic and artificial -- e.g., ``\textit{Box~0 contains the painting, Box~1 contains the bell, Box~2 contains the guitar, \ldots\ Move the glass from Box~6 to Box~4. \ldots\ }'' \citep{KimSchuster2023,Kimetal2024} -- making these tasks closer to solving a reasoning puzzle or executing a program than comprehending natural language. 
The parameter and data requirements reported by these studies may thus reflect demands of the task, rather than limits on entity tracking itself. 
% naturalistic narrative comprehension.
% It remains unclear whether the entity tracking abilities (or failures) observed under such conditions generalise to the kind of situation-model updating that occurs during natural language understanding.

In this work, we address both limitations. 
We developed a controlled experimental paradigm in which short narratives -- programmatically generated but naturalistic and readable -- describe scenes at five levels of situational complexity. 
Entity tracking is assessed via both explicit (generation-based) and implicit (forced-choice or probability read-out) methods, enabling comparison between humans and models under matched conditions, with different task demands \cite{Hu2024Auxiliary}. 
We evaluate two fully open model families across scales -- Pythia (70M--12B) and OLMo~2 (1B--32B) -- alongside larger,  contemporary models (Llama~3.3, Qwen~2.5), and compare their performance to humans ($N = 48$) on the same stimuli. 
To assess whether model performance reflects genuine tracking or reliance on distributional priors, we additionally test all models on narratives containing pseudowords and semantically improbable objects. 

We report four findings. First, human entity tracking degrades specifically with situational complexity, %--not with narrative length: accuracy drops from ${\sim}$88\% at the simplest level to ${\sim}$65\% at the most complex, while recency of mention has no effect, 
revealing a cognitive complexity cost to maintaining structured situation models. 
Second, entity tracking in LMs emerges at far smaller scales than previously reported: robust, human-level accuracy is already present at 410 million parameters, and the ability improves predictably with scale, with the effect of complexity diminishing as models grow, and completely disappearing for contemporary models at 70B scale.  
Third, instruction tuning selectively improves explicit but not implicit entity tracking, indicating that the underlying representational capacity is present in base models prior to alignment.
Fourth, performance is robust across standard, pseudoword, and semantically improbable object types, indicating that models robustly track entities through discourse structure rather than lexical association. 

Together, these results demonstrate that entity tracking -- a key component of language comprehension -- emerges with model scale, at parameter counts far below those previously associated with this ability, and that at sufficient scale language models exceed human accuracy on this task.

Beyond the empirical findings, we share our evaluation paradigm, enabling researchers to study naturalistic entity tracking in small, fully open base models without task-specific fine-tuning.\footnote{Code available at \url{https://github.com/karolinadrozdz/entity_tracking_EMNLP}.}

\section{Related Work}
%A body of work studied whether generative models build structured internal  representations of the domains they are trained on -- so-called ``world models'' -- in artificial domains where the ground-truth is fully specified, such as games or planetary motion. 

Whether generative models build structured internal representations of the domains they are trained on -- so-called ``world models'' -- has been studied extensively in artificial domains where the ground truth is fully specified, such as games or planetary motion. Overall, evidence for such representations has been inconsistent across domains.
On the one hand, models trained on Othello 
move sequences develop internal board-state representations 
that support predictions even on unreachable game states 
\citep{li2022emergent, nanda2023emergent}. 
On the other hand, Transformers trained on 
urban navigation harboured spatially incoherent internal maps 
despite near-perfect prediction \cite{vafa2024evaluating}; and models trained on planetary trajectories failed to 
recover the true, underlying Newtonian mechanics \cite{vafa2025has}. 
Together, in such closed domains, perfect next-token prediction does not imply perfect internal models.  

For natural language, the question of whether language models construct coherent `situation models' and dynamically track entity states has been approached at the representational and mechanistic level. 
\citet{Prakashetal2024}, focusing on a simplified version of the entity tracking task without any state changes, identified circuits implementing entity tracking in base language models and showed that fine-tuning enhances these pathways. 
Studying Transformers that were specifically trained on permutation composition, \citet{li2025language} identified structured parallel algorithms (associative scans) for entity tracking. % with the associative variant generalising better to novel sequence lengths. 
This establishes that the computational machinery exists in Transformers, but not whether this is used in language comprehension or if it emerges from natural language pretraining. 
 
The most direct prior assessment comes from \citet{KimSchuster2023}, who used the previously described boxes-and-objects task and found that pure-text models up to 175 billion parameters failed entirely; tracking first surfaced only in GPT-3.5, a closed, code-specialised model. \citet{Kimetal2024} extended this to open-weight model families, locating robust tracking at 13B parameters with 500 billion additional code tokens, or at 7B only under far more extensive code training (2 trillion tokens). 
However, both evaluate models on this artificial, programmatic task and lack comparison to humans -- leaving open whether their findings generalise to the kind of situation-model construction that characterises natural narrative comprehension. 
Our work addresses both limitations.

\section{Methods}

\subsection{Task Design}

We developed a novel entity tracking task in which participants and models track the locations of objects as they move across locations within short narratives of varying complexity. Each narrative is followed by a question that probes the final location of a target object (see \hyperref[sec:appendix_narratives]{Appendix \ref*{sec:appendix_narratives}} for narrative examples).

\begin{figure}[ht]
  \centering
  \includegraphics[width=\linewidth]{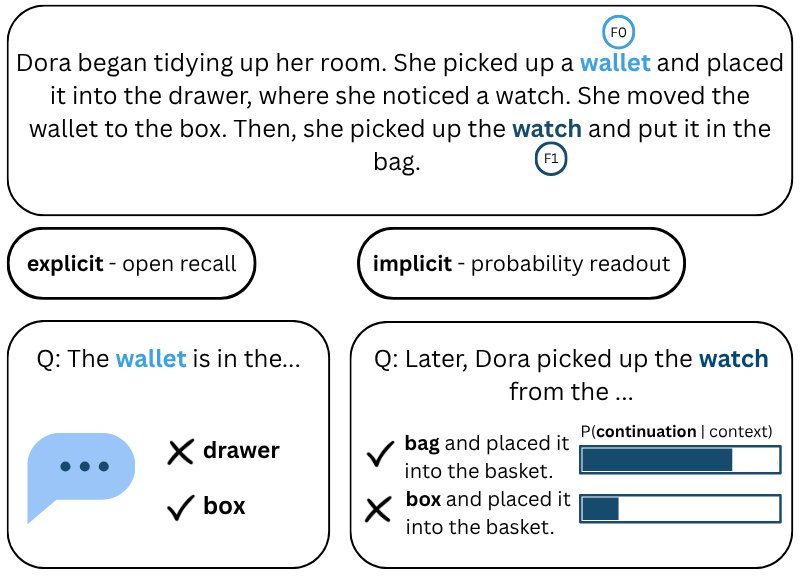}
  \caption{Overview of the experimental design. The top panel displays a simplified narrative at complexity 3, with two target objects (F0, F1). The bottom panels show the two evaluation formats: an \textit{explicit} open recall task (left) requiring generation of the correct final location of the evaluated object, and an \textit{implicit} task (right) administered as a forced-choice selection for human participants and as a probability read-out comparing correct versus incorrect continuations for language models.
  For full example narratives at multiple complexity levels, see \hyperref[sec:appendix_narratives]{Appendix \ref*{sec:appendix_narratives}.}}
  \label{fig:methods}
\end{figure}

\textbf{Complexity.} Complexity was defined by the number of objects, locations, and movements to be tracked. Five levels were constructed: C1 involved a single object in one location; C2 a single object moving between two locations; C3 two objects across three locations; C4 three objects with interleaved movements across three locations; and C5 four objects across four locations. Human participants were tested on C1--C4, as pilot data indicated that C4 already imposed substantial cognitive load; LMs were additionally tested on C5.  

\textbf{Evaluation formats.} Because explicit question-answering can systematically underestimate model capabilities \citep{Hu2024Auxiliary}, we used both explicit and implicit evaluation formats.  
For models, the \textit{explicit} condition required free generation, while the \textit{implicit} condition utilised direct probability read-outs. Correspondingly, for human participants, the \textit{explicit} and \textit{implicit} conditions consisted of open-ended generation and forced-choice selection, respectively (see \hyperref[fig:methods]{Figure \ref*{fig:methods}}).  Unlike in the boxes-and-objects paradigm \citep{KimSchuster2023, Kimetal2024}, in both conditions we ask where the queried object ends up, not what a given location contains.

\textbf{Recency control.} 
Since narratives of higher complexity are also longer, this introduces a potential confound:  accuracy decreases could reflect recency effects rather than complexity \textit{per se}. 
To address this, we varied the position of the target object within longer narratives (i.e., focus object, F0--F3). This allowed us to assess whether performance depended on how recently an object was mentioned, and thus to separate recency-based recall from the demands of tracking multiple entities. If performance were driven by recency, more recently mentioned objects should yield higher accuracy; conversely, stable performance across positions would indicate reliance on integrated representations rather than surface-level recall.

\subsection{Stimuli}

%OLD version

%All narratives were generated using an automated pipeline. 
%This enabled a systematic and scalable construction of novel stimuli that are guaranteed to not appear verbatim in the training data. The resulting narratives were designed to be naturalistic rather than rigid or artificial, following a fixed event structure to maintain coherence. Objects, locations, and characters were randomly sampled for each instance, ensuring diversity across trials.

%NEW - more details:
All narratives were generated using an automated template-based pipeline. 
This enabled a systematic and scalable
construction of novel stimuli that are guaranteed
to not appear verbatim in the training data. 
The resulting narratives were designed to be naturalistic
rather than rigid or artificial, following a fixed event
structure to maintain coherence. 
For each instance, character names, objects, and locations were randomly sampled from fixed pools (6 names, 12 objects, 4 locations) and inserted into the templates, ensuring diversity across trials. We restricted state changes to spatial transfers (picking up, placing, moving between locations), as these yield unambiguous ground-truth object locations; other state changes (e.g., transformations) were outside the scope of this evaluation.

To discourage participants from adopting a strategic search for object locations -- rather than constructing a coherent situation model -- we introduced minor lexical variation (e.g., descriptive adjectives in $\sim$30\% of sentences) and interleaved 10 filler trials with structurally distinct narratives and comprehension questions targeting general story details. 

\textbf{Template variants.} To assess reliance on semantic priors, we constructed three template conditions: (i) \textit{standard}, using objects commonly found in the described context; (ii) \textit{pseudoword}, using pronounceable nonwords generated with Wuggy \cite{keuleers2010wuggy} matched in form; and (iii) \textit{improbable}, using semantically incongruent or abstract entities \cite{wang2018modeling} (see \hyperref[sec:appendix_objects]{Appendix \ref*{sec:appendix_objects}}). Since the confound of exploiting training data distributions is unique to LMs, they require adversarial evaluations to demonstrate robust tracking. Therefore, while the human baseline was established using only the standard template, models were evaluated across all three templates to test whether their performance reflects a generalisable mechanism or merely semantic plausibility.

\subsection{Human Evaluation}
\textbf{Sample.} We recruited $N = 48$ native English speakers via Prolific and compensated them for their participation (mean age = 40.8, $SD = 12.9$; 24 male, 20 female, 4 not disclosed). The data collection protocol was approved by the Ethics Review Board of the Faculty of Social and Behavioural Sciences at the University of Amsterdam (decision number: FMG-8670), and participants provided informed consent. Participants were required to complete the full study and confirm attentiveness and absence of AI assistance. 
Participants with accuracy below 50\% on a given task were excluded from that task’s analysis. 
This yielded final samples of $N = 41$ for the explicit task and $N = 43$ for the implicit task.

\textbf{Procedure.} Participants completed 46 trials (36 main, 10 filler). The main trials were evenly split between explicit and implicit tasks (18 trials each) and balanced across seven configurations (8 $\times$ C1.F0, 8 $\times$ C2.F0, 4 $\times$ C3.F0, 4 $\times$ C3.F1, 4 $\times$ C4.F0, 4 $\times$ C4.F1, 4 $\times$ C4.F2). 
Narratives were presented on a single screen, followed by a probe. 
Reading time was limited to 40 seconds and response time was limited to 15 seconds (including reading the question). 
These constraints were designed to target entity tracking as it naturally occurs during comprehension, rather than treating the task as a reasoning puzzle. 
Importantly, they also greatly reduced the feasibility of external aids such as note-taking or querying LLMs. 

\textbf{Data Analysis.} We fit a logistic generalised linear model with trial-level binary accuracy as the dependent variable. Predictors included mean-centred complexity, probing method, and their interaction, with standard errors clustered by participant. Recency was included as a covariate.
%Recency effects were analysed separately via object position at C4, by fitting logistic generalised linear models predicting accuracy from object slot, with standard errors clustered by participant.

\subsection{Model Evaluation}

\textbf{Model Selection.} We evaluated two fully open model families: Pythia \citep{biderman2023pythia} and OLMo 2 \citep{olmo20252olmo2furious}. These models provide full access to training data and checkpoints, enabling controlled scaling analyses. For the Pythia suite, we tested all available sizes: 70M, 160M, 410M, 1B, 1.4B, 2.8B, 6.9B, and 12B parameters. OLMo 2 models included 1B, 7B, 13B, and 32B variants (base and instruction-tuned). We additionally evaluated two large contemporary models, Llama 3.3 70B \cite{grattafiori2024llama3herdmodels} and Qwen 2.5 72B \cite{qwen2024qwen25}, as high-performing reference models.

\textbf{Evaluation protocol.} Each model was evaluated on 50 trials per condition (complexity $\times$ object position), yielding 550 trials per task. For implicit trials, we computed summed log-probabilities of candidate continuations %using Minicons \citep{misra2022minicons},
scoring a response as correct if the correct continuation received higher probability. 

\begin{figure}[htbp]
  \centering
  \includegraphics[width=\columnwidth]{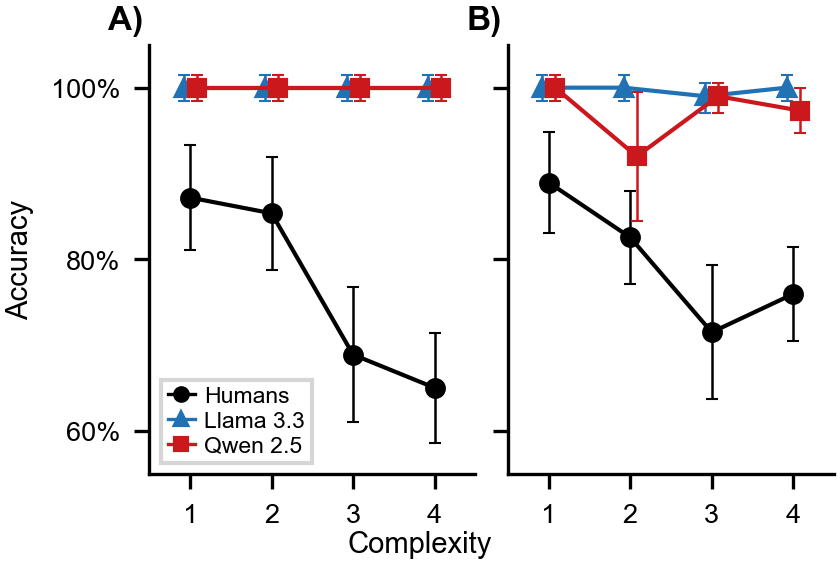}
  \caption{Humans show declining accuracy with increasing complexity while contemporary 70B LMs (Llama 3.3 and Qwen 2.5) maintain ceiling performance. 
  (A) Explicit entity tracking task. (B) Implicit entity tracking task. Error bars represent 95\% confidence intervals.}
  \label{fig:entity_tracking}
\end{figure}

For explicit trials, models were prompted (the exact prompt used can be found in \hyperref[sec:appendix_prompt]{Appendix \ref*{sec:appendix_prompt}}) to produce a single-word location response. Responses were considered correct if they contained the target location term, allowing minor formatting variation. Base models were evaluated primarily on the implicit task, as they are not optimised for instruction-following. 

\textbf{Data Analysis.} Models were treated as individual case studies. Accuracy was computed across conditions, with comparisons focusing on systematic variation across model size, instruction tuning, and template type. Human--model comparisons were conducted descriptively by contrasting the distribution of human accuracy with the performance of individual models under matched conditions.

\section{Results}
\subsection{Human entity tracking performance declines with narrative complexity, independent of recency}

We first asked whether humans reliably track entities at various complexity levels. 
Participants performed well above chance, at 75.3\% accuracy on the explicit and 79.3\% on the implicit task (see \hyperref[sec:appendix_individual]{Appendix~\ref*{sec:appendix_individual}}).

%### FIGURE
\begin{figure}[htbp]
  \centering
  \includegraphics[width=\columnwidth]{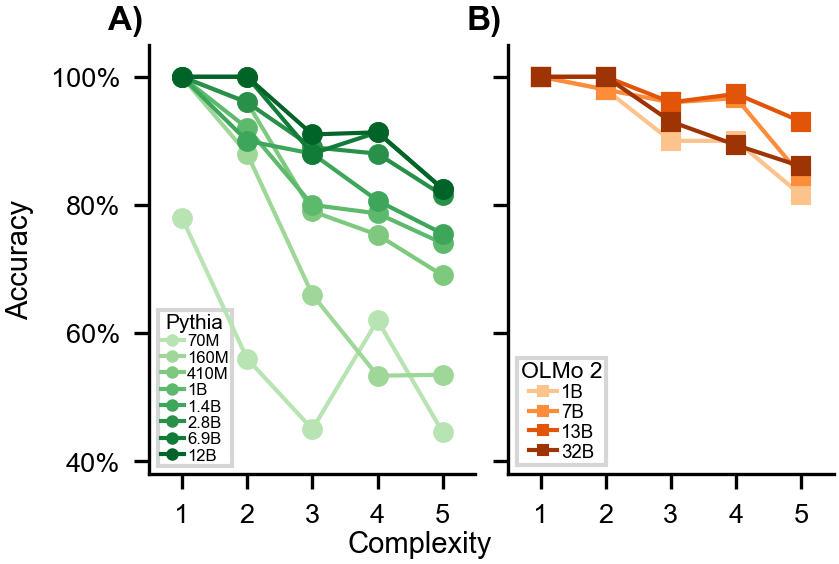}
  \caption{Model scaling effects on implicit entity tracking accuracy. (A) Pythia models (70M to 12B parameters). (B) OLMo 2 base models (1B to 32B parameters). Larger models generally show improved performance, with the exception of OLMo 32B-base.}
  \label{fig:model_scaling}
\end{figure}
%######## 

Notably, performance declined with increasing narrative complexity: in the explicit task, accuracy decreased from 87.2\% at C1 to 65.0\% at C4; in the implicit task, from 89.0\% to 76.0\% ($b = -0.38$, $\text{SE} = 0.07$, $z = -5.72$, $p < 0.001$; \autoref{fig:entity_tracking}); 
each unit increase in complexity reduced the odds of a correct response by approximately 32\% ($\text{OR} = 0.68$, 95\% CI $[0.60, 0.78]$), and this effect was statistically indistinguishable across explicit and implicit tasks ($b = -0.09$, $\text{SE} = 0.05$, $z = -1.73$, $p = 0.08$).

This complexity effect could, in principle, reflect recency or narrative length, since higher-complexity narratives are also longer.
To control for this, our design varied the relevant object's position within longer narratives, allowing us to isolate the effect of recency at matched complexity. Including target object's position as a covariate, the complexity effect remained significant ($b = -0.53$, $SE = 0.10$, $z = -5.45$, $p < 0.001$; $OR = 0.59$, 95\% CI $[0.48, 0.71]$), while recency was not a significant predictor ($b = 0.19$, $SE = 0.11$, $z = 1.63$, $p = 0.10$). Human performance thus appears primarily limited by the complexity of the situation model itself -- the number of entities and relations being tracked -- not by surface-level recency. 

\subsection{Contemporary models maintain ceiling performance across complexity levels}
We then compared human performance against two contemporary 70B-class models, Llama 3.3 and Qwen 2.5. 
While human accuracy fell steeply with complexity, both models held near-ceiling accuracy across all levels (see \autoref{fig:entity_tracking}). 
This model class thus operates far beyond human cognitive capacity, with no trace of the complexity cost that limits human comprehenders. 

% % In stark contrast to humans, contemporary LMs exhibited no complexity effect. Both Llama 3.3 (70B) and Qwen 2.5 (72B) achieved near-perfect accuracy that remained stable across complexity levels (see \autoref{fig:entity_tracking}). 
% This dissociation suggests that the cognitive bottleneck limiting human entity tracking does not constrain these large-scale language models.

\subsection{Entity tracking scales with model size}

Because contemporary models saturate the task, we turned to fully open model suites at smaller scales, to trace where this representational capacity emerges, in both base and instruction-tuned models.  
We evaluated Pythia (70M to 12B parameters) and OLMo 2 (1B to 32B parameters) on the same stimuli (see \autoref{fig:model_scaling}), on both the implicit (forced-choice) and explicit (generation) tasks.

% First, evaluating the base models on the implicit task, we found tracking was present all models and improved predictably with model scale (. 
% Pythia accuracy improved monotonically from 53.5\% (70M) to 89.6\% (12B). 
% OLMo~2 established a higher baseline (88.5\% at 1B; 96.0\% at 13B) with the same pattern of diminishing complexity costs. 
% One exception was OLMo~2 32B, which underperformed the 13B model (90.7\% vs 96.0\% overall); this might reflect training differences in  OLMo~2 32B specifically (see below). %  likely reflects documented architectural and training pipeline modifications at the 32B scale \cite{allenai2025olmo2_32b}, rather than a reversal of the general scaling trend.

On the implicit task, all base models showed substantial entity tracking and exhibited a human-like effect of complexity: performance declined as situational complexity increased (see \autoref{fig:model_scaling}). 
Critically, for both model families, overall performance increased with scale. 
Pythia accuracy improved monotonically from 53.5\% at 70M to 89.6\% at 12B. %, with the C1-to-C5 drop shrinking from 34 points (70M) to 18 points (12B). 
OLMo~2 showed a similar but higher-baseline trajectory, %improving from 88.5\% at 1B to 96.0\% at 13B, 
with smaller complexity costs throughout (e.g., 18 points at 1B; 7 points at 13B). This scaling pattern was statistically robust: logistic regressions predicting accuracy from log(parameters) yielded significant effects for both Pythia ($b = 0.43$, $z = 17.82$, $p < 0.001$, $OR = 1.53$) and OLMo 2 ($b = 0.36$, $z = 4.37$, $p < 0.001$, $OR = 1.43$; excluding the 32B outlier, as detailed below).

One exception to the overall scaling trend was OLMo~2 32B, which underperformed the 13B model. 
We believe this reflects the outlying nature of this particular model -- potentially related to documented modifications to the training pipeline of the 32B model \cite{allenai2025olmo2_32b} -- rather than a genuine reversal of the scaling trend.

On the explicit task, instruction-tuned OLMo~2 models -- which were explicitly trained for question answering -- showed a similar scaling pattern, improving from 55.3\% at 1B to 93.5\% at 32B (%\autoref{fig:instruction_tuning}A;
\hyperref[sec:appendix_scaling]{Appendix~\ref*{sec:appendix_scaling}}). 
Here, the 32B model was again an outlier, but now in a positive sense, far outperforming the 13B (93.5\% vs 65.6\%). Pythia has no instruction-tuned variant, and base models predictably performed at or below chance on the explicit task; thus, we did not analyse their explicit task performance further.

%For completeness, we evaluated the base Pythia models on the explicit task; while numerical, yet not entirely consistent,improvements were observed with scale, overall performance was poor (\hyperref[sec:appendix_scaling]{Appendix~\ref*{sec:appendix_scaling}}), which is expected given that these models were not trained for open-ended question answering.

%not sure where to add it for now 
%Interestingly, neither Pythia nor OLMo 2 base models exhibited a recency effect evaluated at C4: accuracy did not increase for more recently mentioned objects in the implicit task (Pythia: F0: 80.0\%, F1: 75.8\%, F2: 77.0\%; OLMo 2: F0: 95.5\%, F1: 90.5\%, F2: 94.0\%) or the explicit task (Pythia: F0: 23.5\%, F1: 26.2\%, F2: 28.2\%; OLMo 2: F0: 74.0\%, F1: 26.5\%, F2: 71.5\%)

Together, these results indicate that entity tracking is present in base language models below 1 billion parameters, improves predictably with scale, and -- like in humans -- is modulated by situational complexity.

\subsection{Instruction tuning improves explicit but not implicit performance}

\begin{figure}[ht]
  \centering
  \includegraphics[width=\columnwidth]{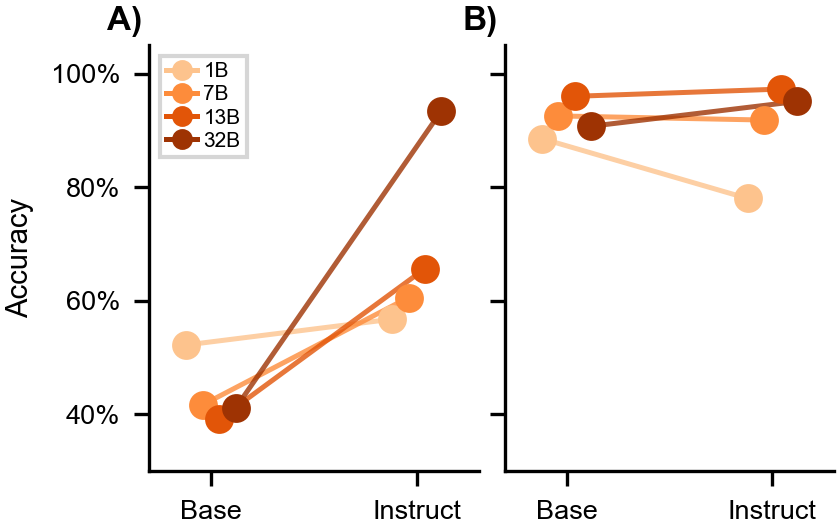}
  \caption{Instruction tuning effects on OLMo 2 models (A) Explicit task. (B) Implicit task. Instruction tuning substantially improves explicit task performance, with gains increasing with model size. In contrast, instruction tuning has mixed effects on implicit task performance.}
  \label{fig:instruction_tuning}
\end{figure}

\begin{figure}[ht]
  \centering
  \includegraphics[width=\columnwidth]{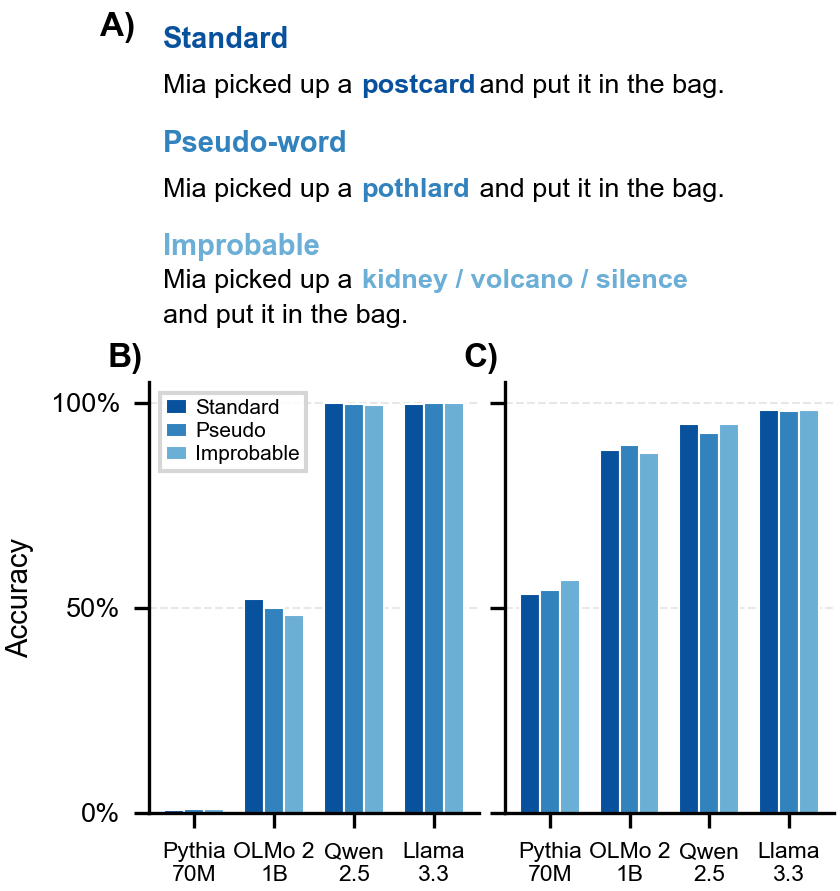}
  \caption{(A) Examples of standard, pseudoword, and improbable object types used in the stimuli. (B) Explicit task accuracy. (C) Implicit task accuracy. Object type does not affect model performance.}
  \label{fig:template_type}
\end{figure}

To examine the effect of instruction tuning, we compared OLMo 2 base models with their instruction-tuned counterparts (see \autoref{fig:instruction_tuning}). 
Instruction tuning substantially improved explicit task performance -- with gains up to +52.4 percentage points -- consistent with the task's demand for open-ended question answering. In contrast, instruction tuning did not improve implicit performance, and in some cases slightly reduced it (e.g., OLMo~2 1B: 88.5\% to 78.0\%). This indicates  instruction tuning enhances the ability to express entity tracking in a question-answering format, without deepening the underlying capability.

\subsection{Entity tracking is robust to semantic content} 

A distinguishing feature of our design is the use of naturalistic narratives rather than the procedural formats of prior work \citep{KimSchuster2023,Kimetal2024}. 
While this makes the task more representative of natural comprehension, it also raises the possibility that models exploit lexical associations -- e.g., inferring a key is in a box because keys and boxes co-occur -- rather than tracking entity movements. 
To control for this, we evaluated all models on three template conditions: standard objects, pseudowords, and semantically improbable objects (\autoref{fig:template_type}; panel A).

Performance was consistent across conditions (\autoref{fig:template_type}; panels B,C). 
Accuracy remained within 3 percentage points of the standard template across all model families and evaluation formats. 
As a manipulation check, we compared sequence log-probabilities across templates and confirmed that models did register the lexical manipulation: pseudoword narratives were $10^{17}$ to $10^{27}$ times less likely than standard narratives, and improbable object narratives were $10^{8}$ to $10^{16}$ times less likely across all evaluated models.
The fact that accuracy was unaffected despite this increased surprisal indicates that entity tracking operates over discourse structure rather than lexical familiarity.

\section{Discussion}

We propose a new, controlled  paradigm for evaluating entity tracking via naturalistic narratives. Our findings challenge and refine key conclusions from prior work on entity tracking in language models.

The most influential prior evaluation of entity tracking  --  the boxes paradigm \citep{KimSchuster2023, Kimetal2024}  --  requires tracking objects through sequences of movements along abstract locations (Box 1, Box 2, etc). 
For humans, solving this task requires step-by-step reasoning and would be difficult without extended time and external aids such as pen and paper. 
This is fundamentally different from the implicit, on-the-fly updating that characterises entity tracking in natural language comprehension. 
Our paradigm targets this latter process using naturalistic narratives and speeded responses. 
Even under this more intuitive task, however, we observe that human entity tracking is far from perfect, underscoring the importance of measuring humans when evaluating cognitive abilities of language models \cite{ivanova2025how}. 

Importantly, in this more naturalistic paradigm, we observe that language models perform far better than previously reported. 
On the implicit task  --  which, especially for smaller models, provides a more sensitive measure of underlying capacities than explicit generation \citep{Hu2024Auxiliary}  --  human-level entity tracking is already present at 410M parameters. 
This stands in stark contrast to prior claims \cite{Kimetal2024,KimSchuster2023} that it requires at least 13B parameters with 500B additional code specialised tokens (Code Llama), or 7B parameters with 2T code tokens (DeepSeek Coder). 
Pythia~410M is over an order of magnitude smaller in parameter scale than the smallest code-augmented model previously found sufficient, and trained on just 300B tokens of predominantly natural language. 
This suggests that the scale and data requirements reported by prior work may reflect the procedural demands of their task rather than fundamental limits on entity tracking. 

The fact that previous studies found code pretraining to be crucial for entity tracking \citep{KimSchuster2023, Kimetal2024} aligns with a broader set of findings that code pretraining can improve many non-code downstream abilities \citep{madaan2022language, petty2025does, aryabumicode}. 
However, code related improvements are typically seen in relatively formal domains such as reasoning \citep{madaan2022language, aryabumicode, ruis2025procedural} and semantic parsing tasks \citep{petty2025does} that differ substantially from naturalistic language comprehension. 
This raises the question whether code pretraining is critical for entity tracking itself, or specifically for the procedural, program-like operationalisation used in prior studies. 
Our results cannot conclusively answer this (see Limitations), but at minimum demonstrate that the requirements for entity tracking depend on how it is measured  --  and that for naturalistic texts, the bar appears lower than previously thought.

\section*{Limitations}
There are multiple limitations to consider when drawing conclusions from this work.

% edited this a bit because we don't really claim this 
First, we only assess entity tracking in an abstracted form,
within short texts.
While entity tracking is a necessary component of discourse comprehension, we do not claim that our results are evidence for human-level discourse comprehension more broadly. 
In fact, even much larger LMs have been shown to fall short of human performance on more subtle, longer-form discourse comprehension tasks, such as plot-hole detection \citep{ahuja2025finding}, indicating that the entity tracking ability we observe here does not immediately extend to all discourse comprehension writ-large. 

Second, we only evaluate existing pretrained models. Although these are fully open, both Pythia \citep[trained on the Pile;][]{biderman2023pythia} and OLMo~2 \citep{olmo20252olmo2furious} include code in their pretraining data, and we therefore cannot isolate the specific contribution of code to the entity tracking capacity we observe.  %cannot make definitive claims about the effect of code training. 
Controlled ablations comparing text-only against code-augmented pretraining, or targeted analyses tracing the influence of specific pretraining data \citep[e.g.,][]{ruis2025procedural}, would be needed to determine whether naturalistic entity tracking emerges without any code exposure.

Third, our output-based evaluation cannot identify the underlying algorithms models use. 
Future mechanistic interpretability work could test whether the entity tracking circuits and algorithms identified in structured tasks \citep{Prakashetal2024, li2025language,tang2026language} also operate during more naturalistic texts, or whether they use different strategies for discourse-based entity tracking, or 
 switch between algorithmic state-tracking and shallower, 'computationally cheaper' heuristics -- analogous to the ``end-game degradation'' observed in game-playing models \cite{nanda2023emergent}. 

Fourth, contemporary models  --  Llama~3.3 \citep{grattafiori2024llama3herdmodels} and Qwen~2.5 \citep{qwen2024qwen25}  --  were included only as high-performing reference points and saturated the task at all tested levels.
Evaluating contemporary models to their breaking point at multiple model scales, and characterising the performance when models (and humans) are allowed to do more extended reasoning, remains an interesting direction for future work.

Fifth, our model-human comparison rests on error rates alone, rather than on the types of errors that models and humans make \cite{geirhos2020beyond}. 
% slight excuse/apology 
This follows from our task design: the implicit task is binary and allows only one error type, while the open response format and unique-vignette setup of our explicit task make errors difficult to group into comparable categories.
Characterising whether humans and models fail in the same ways, and not merely at the same rate, is an important avenue for future work.

Lastly, our stimuli are naturalistic relative to prior procedural tasks \citep{KimSchuster2023, Kimetal2024}, but remain controlled compared to natural discourse. This was necessary to isolate complexity, manipulate recency, and establish ground truth. Extending to longer-form text or dialogue would improve ecological validity at the cost of experimental control.

\section*{Acknowledgements}

This work has been partly funded by the Dutch Research Council (NWO) under Veni grant VI.Veni.231G.043 to MH.

%%%NOTES
%Beckmann --> heuristics vs complex cicruci, system 1 vs system 2 future research!!! " Moreover, the fidelity of this emergent state-of-the-world understanding is not uniform. Nanda et al. observed “end-game degradation,” where the model’s reliance on its internal representation of  the board state seemed to falter in the final moves of a game. They hypothesized that as the board fills  up and options become fewer, it becomes computationally cheaper for the model to switch from a  complex world-modeling circuit to simpler heuristics." 

%"They may pragmatically toggle between modeling the state of the world and  relying on shallow heuristics, depending on which circuit minimizes loss most efficiently. Thus, while  the Othello-GPT findings suggest that LMs can maintain an internal representation of the state of the  external world, they also warn us that they will not necessarily do so when a shortcut is available."

%but 5 complexity levels for models is too little --> Although models could, in principle, handle arbitrarily long and complex narratives, we restricted evaluation to these controlled levels to ensure comparability with human performance. 


\bibliography{custom}

\clearpage
\appendix
\section{Performance of OLMo 2 Instruct Models on the Explicit Task}
\label{sec:appendix_scaling}

\begin{figure}[ht]
  \centering
  \includegraphics[width=0.7\linewidth]{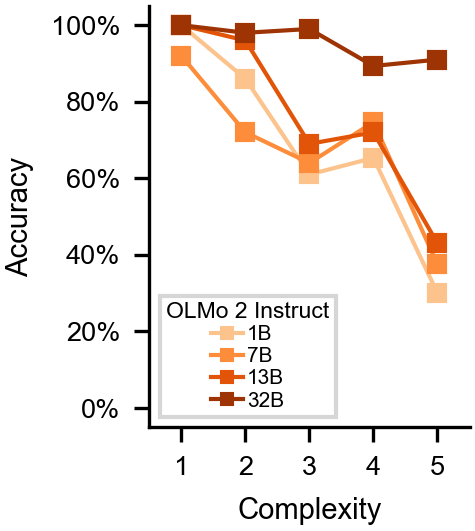}
  \caption{OLMo 2 Instruct models on the explicit (generation) task. Instruction-tuned models show substantial scaling effects, with accuracy improving from 55\% (1B) to 94\% (32B) overall. The 32B model maintains high accuracy even at C5, while smaller instruction-tuned models show steep performance declines with increasing complexity.}
  \label{fig:pythia_olmo_scaling}
\end{figure}
%%
%\section{Performance of Open Base vs. Instruction-Tuned Models on the Explicit Task}
%\label{sec:appendix_scaling}

%\begin{figure}[ht]
  %\centering
  %\includegraphics[width=\linewidth]{figures/figE.png}
  %\caption{(A) Pythia base models (not instruction-tuned) on the explicit task. Performance is poor overall, with smaller models showing some improvement with scale, but larger models (6.9B, 12B) performing no better than mid-sized models (2.8B). (B) OLMo 2 Instruct models on the explicit task. Instruction-tuned models perform substantially better and exhibit a clear scaling pattern, with larger models consistently outperforming smaller ones.}
  %\label{fig:pythia_olmo_scaling}
%\end{figure}

\section{Object Words by Template Condition}
\label{sec:appendix_objects}

\raggedbottom
\begin{table}[H]
  \centering
  \begin{tabular}{lp{4.8cm}}
    \hline
    \textbf{Template} & \textbf{Objects} \\
    \hline
    \textbf{Standard} & ring, key, postcard, pen, bookmark, book, wallet, watch, coin, photo, scarf, letter \\
    \textbf{Pseudoword} & rint, kel, pothlard, pes, boolmarm, bood, gallel, wawed, coft, knolo, scact, lunter \\
    \textbf{Improbable} & \\
    \quad Rare & kidney, squid, brick, urn \\
    \quad Phys.  Impos. & volcano, mountain, skyscraper, cathedral \\
    \quad Abstract & happiness, silence, justice, peace \\
    \hline
  \end{tabular}
  \caption{Object words used in each template condition. The conditions include: 1) common \& possible objects (standard template; $N = 12$), 2) pseudowords derived from the standard objects ($N = 12$), 3a) rare but possible objects ($N = 4$), 3b) rare and physically impossible objects (size violation; $N = 4$), and 3c) rare and logically impossible objects (abstract violation; $N = 4$).}
  \label{tab:template_objects}
\end{table}

\section{Evaluation Prompt}
\label{sec:appendix_prompt}

For the explicit evaluation, we used the following instruction prompt across all models:

\vspace{1em}
\begin{tcolorbox}[colback=gray!10, colframe=gray!50!black, arc=4pt, boxrule=0.5pt, left=6pt, right=6pt, top=6pt, bottom=6pt]
Follow these instructions:

1. Carefully read the story and question to identify the object's final location.

2. Your response must contain ONLY the single word for that final location (e.g., if the final location is 'the red box', respond 'box'; if it's 'the bag', respond 'bag').

3. DO NOT include any reasoning, explanations, introductory phrases, sentences, or punctuation. Your entire response should be just the single location word.

Location:
\end{tcolorbox}
\vspace{1em}

\section{Distribution of Individual Human Performance}
\label{sec:appendix_individual}
\begin{figure}[ht]
  \centering
  \includegraphics[width=0.9\linewidth]{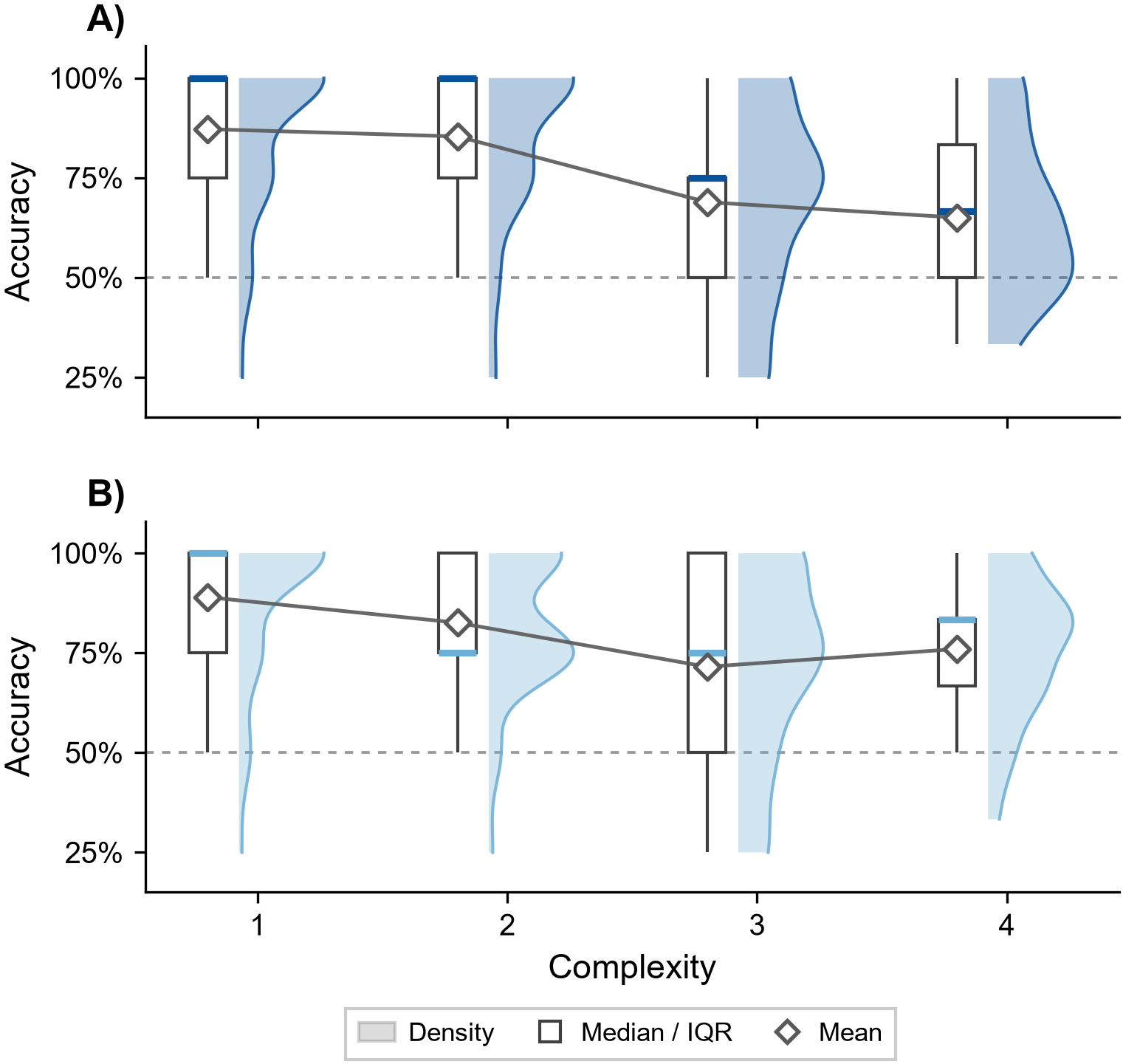}
  \caption{Distribution of human accuracy across complexity levels on the explicit (A; $N = 41$) and implicit (B; $N = 43$) tasks. Half-violins show kernel density estimates of the between-participant distribution, boxes show the median and interquartile range, diamonds indicate group means.}
  \label{fig:individual_performance}
\end{figure}

\section{Example Narratives}
\label{sec:appendix_narratives}

\noindent \textbf{Explicit Task} 
\vspace{1em}

\noindent \textit{Complexity 1, Focus Object 0}

\noindent Dora began tidying up her room. She picked up a book and put it into the basket. She paused to check her phone. Q: The book is in the ...      

\noindent \textit{Expected answer}: basket
\vspace{1em}

\noindent \textit{Complexity 2, Focus Object 0}

\noindent Lila began tidying up her room. She picked up a bookmark and placed it into the bag. She moved the bookmark from the bag to the drawer. She glanced out the window. Q: The bookmark is in the ...

\noindent \textit{Expected answer}: drawer
\vspace{1em}

\noindent \textit{Complexity 3, Focus Object 0}

\noindent Sophia began tidying up her room. She picked up a key and carefully placed it into the blue basket, where she noticed a photo. Curious, she moved the key from the blue basket to the modern box. Then, she moved the photo from the blue basket to the vintage bag. She glanced out the window. Q: The key is in the ...

\noindent \textit{Expected answer}: box
\vspace{1em}

\noindent \textit{Complexity 3, Focus Object 1}

\noindent Lila began tidying up her room. She picked up a postcard and carefully placed it into the vintage drawer, where she noticed a letter. Curious, she moved the postcard from the vintage drawer to the box. Then, she moved the letter from the vintage drawer to the bag. She took a sip of water. Q: The letter is in the ...

\noindent \textit{Expected answer}: bag
\vspace{1em}

\noindent \textit{Complexity 4, Focus Object 0}

\noindent Anna began tidying up her room. She picked up a postcard and carefully placed it into the box. While organizing, she noticed a scarf in the basket. Curious, she moved the scarf to the vintage drawer and found a watch there. She put the watch into the box. She paused to check her phone. Q: The postcard is in the ...

\noindent \textit{Expected answer}: box
\vspace{1em}

\noindent \textit{Complexity 4, Focus Object 1}

\noindent Mia began tidying up her room. She picked up a pen and carefully placed it into the small box. While organizing, she noticed a coin in the colorful drawer. Curious, she moved the coin to the colorful basket and found a wallet there. She put the wallet into the small box. She paused for a moment to stretch. Q: The coin is in the ...

\noindent \textit{Expected answer}: basket
\vspace{1em}

\noindent \textit{Complexity 4, Focus Object 2}

\noindent Emma began tidying up her room. She picked up a photo and carefully placed it into the basket. While organizing, she noticed a pen in the box. Curious, she moved the pen to the drawer and found a watch there. She put the watch into the basket. She paused for a moment to stretch. Q: The watch is in the ...

\noindent \textit{Expected answer}: basket
\vspace{1em}

\noindent \textit{Complexity 5, Focus Object 0}

\noindent Sophia began tidying up her room. She picked up a pen and carefully placed it into the bag. While organizing, she noticed a ring in the box. Curious, she moved the ring to the drawer and found a photo there. She picked up the photo and put it into the basket. Then, Sophia moved the scarf from the basket to the box. She paused for a moment to stretch. Q: The pen is in the ...

\noindent \textit{Expected answer}: bag
\vspace{1em}

\noindent \textit{Complexity 5, Focus Object 1}

\noindent Lila began tidying up her room. She picked up a book and carefully placed it into the modern basket. While organizing, she noticed a letter in the drawer. Curious, she moved the letter to the bag and found a coin there. She picked up the coin and put it into the box. Then, Lila moved the photo from the box to the drawer. She took a sip of water. Q: The letter is in the ...

\noindent \textit{Expected answer}: bag
\vspace{1em}

\noindent \textit{Complexity 5, Focus Object 2}

\noindent Anna began tidying up her room. She picked up a bookmark and carefully placed it into the basket. While organizing, she noticed a letter in the bag. Curious, she moved the letter to the vintage drawer and found a key there. She picked up the key and put it into the modern box. Then, Anna moved the ring from the modern box to the bag. She paused to check her phone. Q: The key is in the ...

\noindent \textit{Expected answer}: box
\vspace{1em}

\noindent \textit{Complexity 5, Focus Object 3}

\noindent Emma began tidying up her room. She picked up a wallet and carefully placed it into the bag. While organizing, she noticed a key in the vintage box. Curious, she moved the key to the drawer and found a watch there. She picked up the watch and put it into the basket. Then, Emma moved the postcard from the basket to the vintage box. She paused for a moment to stretch. Q: The postcard is in the ...

\noindent \textit{Expected answer}: box
\vspace{2em}

\noindent \textbf{Implicit Task}
\vspace{1em}

\noindent \textit{Complexity 1, Focus Object 0}

\noindent \textit{Correct continuation}: \\
\noindent Dora began tidying up her room. She picked up a photo and put it into the modern bag. She glanced out the window. Later, Dora picked up the photo from the bag and placed it into the drawer.

\vspace{0.5em}
\noindent \textit{Incorrect continuation}: \\
\noindent Dora began tidying up her room. She picked up a photo and put it into the modern bag. She glanced out the window. Later, Dora picked up the photo from the drawer and placed it into the box.
\vspace{1em}

\noindent \textit{Complexity 2, Focus Object 0}

\noindent \textit{Correct continuation}: \\
\noindent Emma began tidying up her room. She picked up a wallet and placed it into the bag. She moved the wallet from the bag to the box. She glanced out the window. Later, Emma picked up the wallet from the box and placed it into the bag.

\vspace{0.5em}
\noindent \textit{Incorrect continuation}: \\
\noindent Emma began tidying up her room. She picked up a wallet and placed it into the bag. She moved the wallet from the bag to the box. She glanced out the window. Later, Emma picked up the wallet from the bag and placed it into the basket.
\vspace{1em}

\noindent \textit{Complexity 3, Focus Object 0}

\noindent \textit{Correct continuation}: \\
\noindent Dora began tidying up her room. She picked up a watch and carefully placed it into the small bag, where she noticed a wallet. Curious, she moved the watch from the small bag to the basket. Then, she moved the wallet from the small bag to the box. She paused to check her phone. Later, Dora picked up the watch from the basket and placed it into the bag.

\vspace{0.5em}
\noindent \textit{Incorrect continuation}: \\
\noindent Dora began tidying up her room. She picked up a watch and carefully placed it into the small bag, where she noticed a wallet. Curious, she moved the watch from the small bag to the basket. Then, she moved the wallet from the small bag to the box. She paused to check her phone. Later, Dora picked up the watch from the bag and placed it into the box.
\vspace{1em}

\noindent \textit{Complexity 3, Focus Object 1}

\noindent \textit{Correct continuation}: \\
\noindent Mia began tidying up her room. She picked up a bookmark and carefully placed it into the blue drawer, where she noticed a key. Curious, she moved the bookmark from the blue drawer to the basket. Then, she moved the key from the blue drawer to the bag. She paused to check her phone. Later, Mia picked up the key from the bag and placed it into the box.

\vspace{0.5em}
\noindent \textit{Incorrect continuation}: \\
\noindent Mia began tidying up her room. She picked up a bookmark and carefully placed it into the blue drawer, where she noticed a key. Curious, she moved the bookmark from the blue drawer to the basket. Then, she moved the key from the blue drawer to the bag. She paused to check her phone. Later, Mia picked up the key from the box and placed it into the drawer.
\vspace{1em}

\noindent \textit{Complexity 4, Focus Object 0}

\noindent \textit{Correct continuation}: \\
\noindent Sophia began tidying up her room. She picked up a pen and carefully placed it into the box. While organizing, she noticed a scarf in the drawer. Curious, she moved the scarf to the bag and found a ring there. She put the ring into the box. She glanced out the window. Later, Sophia picked up the pen from the box and placed it into the basket.

\vspace{0.5em}
\noindent \textit{Incorrect continuation}: \\
\noindent Sophia began tidying up her room. She picked up a pen and carefully placed it into the box. While organizing, she noticed a scarf in the drawer. Curious, she moved the scarf to the bag and found a ring there. She put the ring into the box. She glanced out the window. Later, Sophia picked up the pen from the basket and placed it into the drawer.
\vspace{1em}

\noindent \textit{Complexity 4, Focus Object 1}

\noindent \textit{Correct continuation}: \\
\noindent Anna began tidying up her room. She picked up a bookmark and carefully placed it into the small bag. While organizing, she noticed a watch in the basket. Curious, she moved the watch to the drawer and found a letter there. She put the letter into the small bag. She paused for a moment to stretch. Later, Anna picked up the watch from the drawer and placed it into the bag.

\vspace{0.5em}
\noindent \textit{Incorrect continuation}: \\
\noindent Anna began tidying up her room. She picked up a bookmark and carefully placed it into the small bag. While organizing, she noticed a watch in the basket. Curious, she moved the watch to the drawer and found a letter there. She put the letter into the small bag. She paused for a moment to stretch. Later, Anna picked up the watch from the bag and placed it into the basket.
\vspace{1em}

\noindent \textit{Complexity 4, Focus Object 2}

\noindent \textit{Correct continuation}: \\
\noindent Lila began tidying up her room. She picked up a coin and carefully placed it into the small drawer. While organizing, she noticed a book in the basket. Curious, she moved the book to the bag and found a watch there. She put the watch into the small drawer. She paused to check her phone. Later, Lila picked up the watch from the drawer and placed it into the basket.

\vspace{0.5em}
\noindent \textit{Incorrect continuation}: \\
\noindent Lila began tidying up her room. She picked up a coin and carefully placed it into the small drawer. While organizing, she noticed a book in the basket. Curious, she moved the book to the bag and found a watch there. She put the watch into the small drawer. She paused to check her phone. Later, Lila picked up the watch from the basket and placed it into the bag.
\vspace{1em}

\noindent \textit{Complexity 5, Focus Object 0}

\noindent \textit{Correct continuation}: \\
\noindent Lila began tidying up her room. She picked up a scarf and carefully placed it into the blue box. While organizing, she noticed a letter in the red drawer. Curious, she moved the letter to the bag and found a postcard there. She picked up the postcard and put it into the vintage basket. Then, Lila moved the photo from the vintage basket to the red drawer. She paused to check her phone. Later, Lila picked up the scarf from the box and placed it into the basket.

\vspace{0.5em}
\noindent \textit{Incorrect continuation}: \\
\noindent Lila began tidying up her room. She picked up a scarf and carefully placed it into the blue box. While organizing, she noticed a letter in the red drawer. Curious, she moved the letter to the bag and found a postcard there. She picked up the postcard and put it into the vintage basket. Then, Lila moved the photo from the vintage basket to the red drawer. She paused to check her phone. Later, Lila picked up the scarf from the basket and placed it into the drawer.
\vspace{1em}

\noindent \textit{Complexity 5, Focus Object 1}

\noindent \textit{Correct continuation}: \\
\noindent Lila began tidying up her room. She picked up a watch and carefully placed it into the blue basket. While organizing, she noticed a wallet in the box. Curious, she moved the wallet to the drawer and found a key there. She picked up the key and put it into the bag. Then, Lila moved the coin from the bag to the box. She paused to check her phone. Later, Lila picked up the wallet from the drawer and placed it into the bag.      

\vspace{0.5em}
\noindent \textit{Incorrect continuation}: \\
\noindent Lila began tidying up her room. She picked up a watch and carefully placed it into the blue basket. While organizing, she noticed a wallet in the box. Curious, she moved the wallet to the drawer and found a key there. She picked up the key and put it into the bag. Then, Lila moved the coin from the bag to the box. She paused to check her phone. Later, Lila picked up the wallet from the bag and placed it into the box.
\vspace{1em}

\noindent \textit{Complexity 5, Focus Object 2}

\noindent \textit{Correct continuation}: \\
\noindent Mia began tidying up her room. She picked up a coin and carefully placed it into the basket. While organizing, she noticed a watch in the modern box. Curious, she moved the watch to the drawer and found a wallet there. She picked up the wallet and put it into the bag. Then, Mia moved the ring from the bag to the modern box. She paused for a moment to stretch. Later, Mia picked up the wallet from the bag and placed it into the basket.

\vspace{0.5em}
\noindent \textit{Incorrect continuation}: \\
\noindent Mia began tidying up her room. She picked up a coin and carefully placed it into the basket. While organizing, she noticed a watch in the modern box. Curious, she moved the watch to the drawer and found a wallet there. She picked up the wallet and put it into the bag. Then, Mia moved the ring from the bag to the modern box. She paused for a moment to stretch. Later, Mia picked up the wallet from the basket and placed it into the box.
\vspace{1em}

\noindent \textit{Complexity 5, Focus Object 3}

\noindent \textit{Correct continuation}: \\
\noindent Dora began tidying up her room. She picked up a scarf and carefully placed it into the bag. While organizing, she noticed a pen in the drawer. Curious, she moved the pen to the big box and found a ring there. She picked up the ring and put it into the small basket. Then, Dora moved the watch from the small basket to the drawer. She took a sip of water. Later, Dora picked up the watch from the drawer and placed it into the box.

\vspace{0.5em}
\noindent \textit{Incorrect continuation}: \\
\noindent Dora began tidying up her room. She picked up a scarf and carefully placed it into the bag. While organizing, she noticed a pen in the drawer. Curious, she moved the pen to the big box and found a ring there. She picked up the ring and put it into the small basket. Then, Dora moved the watch from the small basket to the drawer. She took a sip of water. Later, Dora picked up the watch from the box and placed it into the basket.

\end{document}